\pdfoutput=1

\documentclass[11pt]{article}

\usepackage{acl}

\usepackage{times}
\usepackage{latexsym}

\usepackage[T1]{fontenc}

\usepackage[utf8]{inputenc}

\usepackage{microtype}
\newcommand*\samethanks[1][\value{footnote}]{\footnotemark[#1]}
\usepackage{graphicx}
\usepackage{float}
\usepackage{subfigure}
\usepackage{multirow}
\usepackage{multicol}
\usepackage{booktabs}

\newcommand{\eat}[1]{}

\title{Towards More Effective and Economic Sparsely-Activated Model}

\author{Hao Jiang\textsuperscript{1}\thanks{{} {} Equal contribution.},~ Ke Zhan\textsuperscript{1}\samethanks,~Jianwei Qu\textsuperscript{1}\samethanks,~ Yongkang Wu\textsuperscript{1}\samethanks,~ Zhaoye Fei\textsuperscript{2}\samethanks,~ Xinyu Zhang\textsuperscript{1}\samethanks \\
~ Lei Chen\textsuperscript{3},~ Zhicheng Dou\textsuperscript{4}, ~ Xipeng Qiu\textsuperscript{2},~ Zikai Guo\textsuperscript{1},~ Ruofei Lai\textsuperscript{1},~ Jiawen Wu\textsuperscript{2},~ Enrui Hu\textsuperscript{1} \\
~ Yinxia Zhang\textsuperscript{1},~ Yantao Jia\textsuperscript{1},~ Fan Yu\textsuperscript{1},~ Zhao Cao\textsuperscript{1}\thanks{{} {} Corresponding author.} \\
\normalsize $^1$ Distributed and Parallel Software Lab, Huawei\\

\normalsize $^2$School of Computer Science, Fudan University\\
\normalsize $^3$Department of Computer Science and Engineering, Hong Kong University of Science and Technology\\
\normalsize $^4$Gaoling School of Artificial Intelligence, Renmin University of China\\
\normalsize \tt \{jianghao66,zhanke2,qujianwei,wuyongkang7,zhangxinyu35,caozhao1\}@huawei.com\\
\normalsize \tt zyfei20@fudan.edu.cn\\

}

\begin{document}
\maketitle

\begin{abstract}

The sparsely-activated models have achieved great success in natural language processing through large-scale parameters and relatively low computational cost, and gradually become a feasible technique for training and implementing extremely large models. Due to the limit of communication cost, activating multiple experts is hardly affordable during training and inference. Therefore, previous work usually activate just one expert at a time to alleviate additional communication cost.
Such routing mechanism limits the upper bound of model performance. In this paper, we first investigate a phenomenon that increasing the number of activated experts can boost the model performance with higher sparse ratio. To increase the number of activated experts without an increase in computational cost, we propose SAM (Switch and Mixture) routing, an efficient hierarchical routing mechanism that activates multiple experts in a same device (GPU). Our methods shed light on the training of extremely large sparse models and experiments prove that our models can achieve significant performance gain with great efficiency improvement.

\end{abstract}

\section{Introduction}

Recently, we have observed the great success of large-scale models in many areas. For example, in natural language processing(NLP), many models~\cite{tom-2020-gpt3, deepak-21-megathronlm, Radford2018ImprovingLU, radford-2019-gpt2, colin-2019-t5, mike-2019-bart} based on Transformer~\cite{ashish-2017-transformer} have been proposed to address the problems existing in natural language understanding~\cite{devlin-2019-bert, yihan-2019-roberta} and natural language generation~\cite{colin-2019-t5, mike-2019-bart}. To make these models more powerful, we need to increase the parameter size ~\cite{tom-2020-gpt3, devlin-2019-bert, jared-2020-scale}. Unfortunately, for dense models, extremely large parameter size always comes with extremely huge computational cost. Thus, to reduce computational cost for large models, the sparse activation mechanism is implemented. In contrast to dense models, sparse models only activate partial parameters for computing. ~\cite{noam-2017-moe} uses a Mixture of Experts (MoE) layer to expand LSTM~\cite{Ho-97-lstm} and trains a giant model up to 137B parameters.~\cite{dmitry-2020-gshard} proposed G-Shard which scales up the Transformer model with Sparsely-Gated Mixture-of-Experts layer. However, because of the implementation mechanism of sparsely activated models and the limitations of current hardware, it inevitably produces the communication cost among the drivers when activating multiple experts. To address this,~\cite{william-2021-swt} propose Switch Transformer which selects one expert for calculation, trains a sparsely activated model based on T5 towards the trillion level, and achieves SOTA results in many NLP tasks.

\begin{figure}
    \centering
    \includegraphics[width=0.55\textwidth]{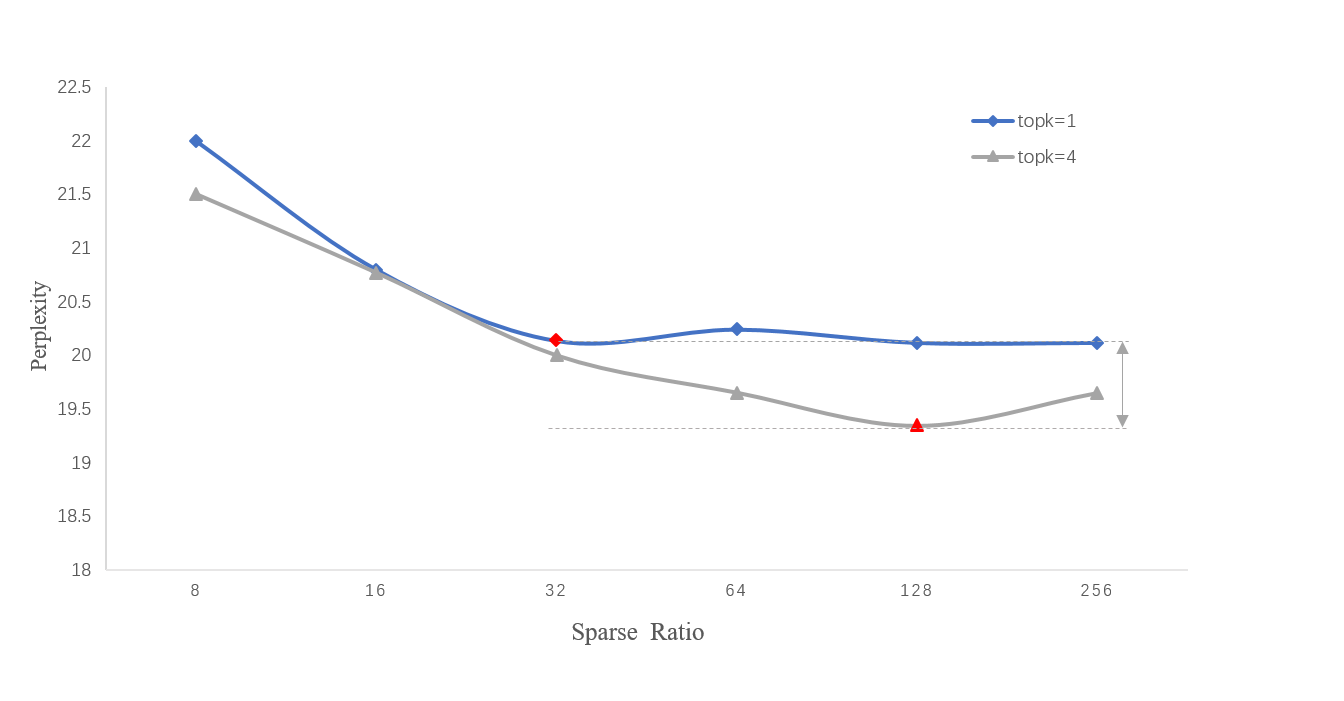}
    \caption{\textbf{Performance with different sparsity ratio} sparsity ratio is $ n_{expert}/k $ which describe Section \ref{topk}. The performance of Switch routing which activate single expert will reach the ceiling when sparsity ratio is 32. When activating 4 experts (also use Switch routing), this ceiling will achieved when sparsity ratio is 128, and the performance is also better than activated single expert.}
    \label{sr_loss}
\end{figure}

Although Switch Transformer~\cite{william-2021-swt} can considerably reduce computational cost, we find that activating only one expert at a time may limit the upper bound of model performance, compared with activating multiple experts. To illustrate this, we introduce a concept of \textit{sparsity ceiling} to denote the optimal performance of sparse model under different sparsity. Let \textit{sparse ratio} denote ratio of the total number of experts to the number of active experts. As shown in Figure \ref{sr_loss}, we can see the performance of model activating single expert (using Switch routing) will get the optimal result when sparsity ratio is 32. Moreover, the model performance cannot be improved further when the sparsity ratio is increased.
Another observation is that  increasing the number of activated experts can improve the optimal sparsity ratio and the sparsity ceiling.

However, for current sparse routing mechanism, increasing the number of activated experts will greatly increase the communication overhead. To address this issue, we propose SAM routing mechanism, which adopts a hierarchical routing to reach the extremely sparsity ratio with acceptable cost. Specifically, we deploy untrained experts in different devices and group them based on their position.
In the forward calculation, the router will activate multiple experts in one expert group. Through gathering in a single device, we decouple the communication overhead with the number of activated experts to achieve efficient and scalable sparsely activated models. Experiments show that SAM routing achieves a significant improvement compared with previous work.

To summarize, our contributions are listed as follows:

\begin{enumerate}
    \item We analyzed the relationship between the sparsity and the performance of sparsely activated models, and found there is a ceiling when increasing the sparsity continuously. Increasing the number of activated experts will break through the ceiling and achieve the extreme sparsity of sparsely activated models.
    \item In order to decouple the communication cost with the number of activated experts, we propose the SAM routing, which implements an efficient and scalability sparse routing mechanism through expert grouping and gathering inner group. Through the SAM routing, we can increase the number of activated experts with a constant communication overhead to increase the scale of the model.


    \item Furthermore, we have designed two losses to align the best experts of local and global, so that the experts we selected through hierarchical routing will be the most suitable ones among all the experts.

    \item The extensive experiments demonstrate that the SAM routing improves the model performance significantly with constant computational cost and communication cost.
\end{enumerate}

\begin{figure}
    \centering
    \includegraphics[width=0.45\textwidth]{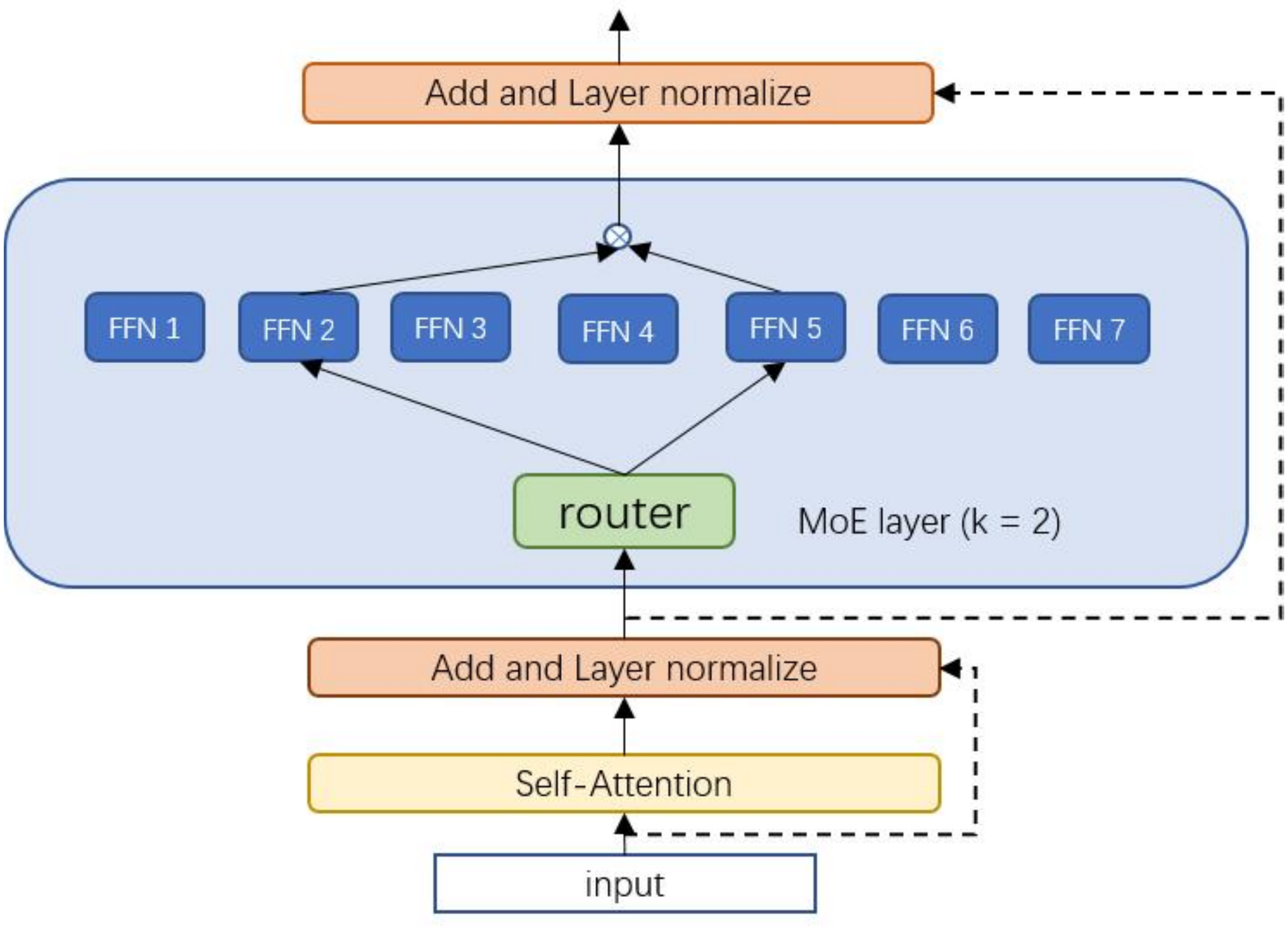}
    \caption{\textbf{The architecture of G-Shard~\cite{dmitry-2020-gshard} using Mixture of Experts (MoE) layer}}
    \label{moe_architecture}
\end{figure}

\section{Background and Analysis}

\subsection{Sparsely Activated Models}

Sparsely activated models are currently considered as very advantageous structures for training huge models. By routing samples to different experts, the model parameter size can be greatly increased and consequently performance can be improved.
G-Shard~\cite{dmitry-2020-gshard} use Mixture of Experts (MoE) layer to train a sparsely activated model based on Transformer~\cite{ashish-2017-transformer}, and the architecture is shown in Figure \ref{moe_architecture}. The token will be routed by router to different experts based on the representation of it.
Specifically, token $x$ will be routed to $i$-th expert according to its score~\cite{noam-2017-moe}:

\begin{equation}
    g(x) =  (w \cdot h_x) + {\rm noise}(h_x),
    \label{router_1}
\end{equation}

\begin{equation}
    p(x)_i =  {\rm softmax}({\rm topk}(g(x)))_i,
    \label{router_2}
\end{equation}

where $ h_x $ is the representation of $x$ in the previous layers, $w$ is learnable parameters in this router and $ p(x)_i $ is the score that $x$ will be routed to $i$-th expert. The Switch Transformer~\cite{william-2021-swt} argues that activating more than one expert will result in unacceptable communication cost, but the router of MoE layer is non-differentiable using only single expert.
Therefore, Switch Transformer simplifies the router
and calculates the score as:

\begin{equation}
    p(x)_i =  {\rm softmax}(w \cdot h_x)_i.
    \label{switch_router}
\end{equation}

For input $x$, the output is determined by the selected expert of the results:

\begin{equation}
    y = \sum_{i \in \mathcal{T}}p(x)_iE(x)_i.
    \label{router_3}
\end{equation}

As shown in the Equation (\ref{router_3}), $ E_i(x) $  represents the output of $i$-th expert, $ \mathcal{T} $  represents the set of index of the selected experts. The final output of the MoE layer is the weighted sum of the output of the selected experts.


\subsection{Expert Parallel}

\begin{figure}
    \centering
    \includegraphics[width=0.45\textwidth]{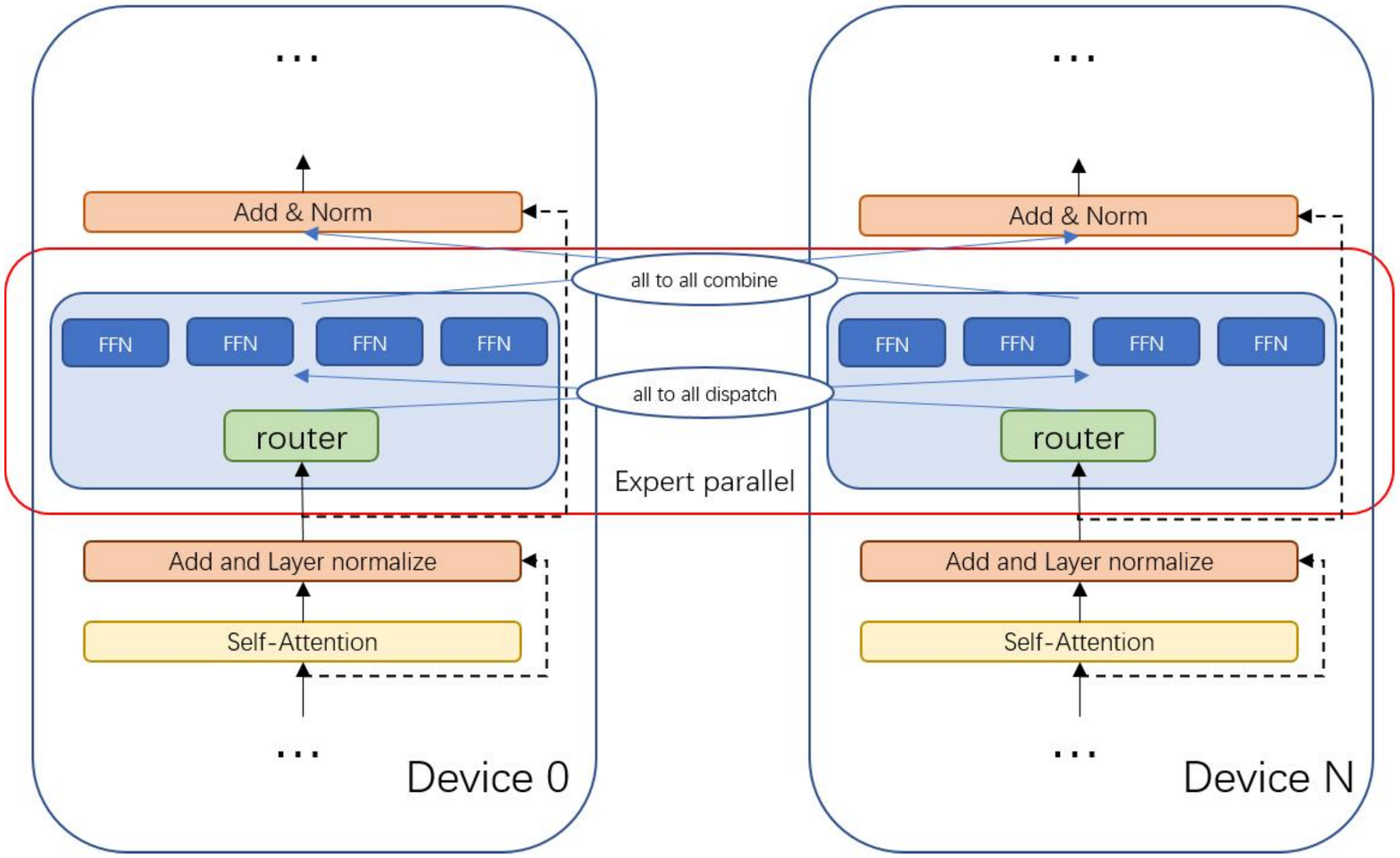}
    \caption{\textbf{the example of expert parallel}, the expert of MoE
     layer will be distributed to multiple devices.}
    \label{expert_parallel}
\end{figure}

Training huge scale model requires distributed training system with high performance and quite some large-scale pre-trained models based on Transformer are successfully trained through parallel training. In addition to data parallelism and model parallelism, expert parallelism, which assigns experts from a single device to multiple devices (generally GPU), is also used when training sparsely activated models, as shown in Figure \ref{expert_parallel}.



As the Figure \ref{expert_parallel} shown, when activating multiple experts, AllToAll operator will cause the communication overhead. This operator will dispatch the token to multiple experts and gather the results after computation. As ~\cite{dmitry-2020-gshard} described, this communication overhead increases sublinearly when the number of device grows.

Switch Transformer~\cite{william-2021-swt}
propose activating only a single expert to avoid this additional communication overhead. However, activating single expert limits the gain obtained through increasing the model capacity, and when the model capacity reaches a certain limit, the model obtains no gains.
The details will be discussed in Section \ref{topk}.

\begin{figure*}
    \centering
    \includegraphics[width=1\textwidth]{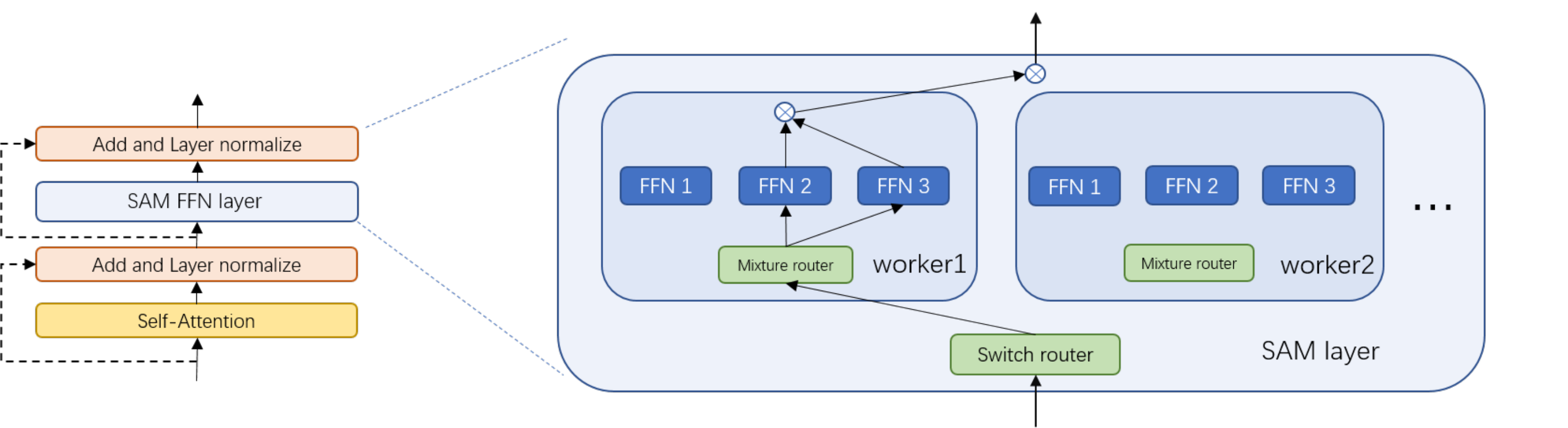}
    \caption{\textbf{The architecture of Switch and Mixture (SAM) layer.}}
    \label{sam_architecture}
\end{figure*}

\subsection{Rethinking Sparsity and Beyond}\label{topk}

According to the model structure of sparsely activated models, one of their advantages is that only part of experts will be activated for forward calculation, which reduces the calculation cost of the models. Previous work \cite{noam-2017-moe, dmitry-2020-gshard, william-2021-swt} has shown that increasing the total number of experts can improve the performance of the model. However, we found \textit{when the number of experts is sufficiently high, no further improvement can be obtained in the sparsely activated models.}


Before we introduce our finding, we first define some notations about the sparsely activated models: $ k $ is the number of activated experts in each layer, $ n_{expert} $ is the number of experts in each layer, $ s_{model} $ is the parameter size of the layer, $ d_{model} $ is the dimension of the model embedding, $ d_{ffn}$ is the dimension of the Feed-Forward Network (FFN) layer. To measure the sparsity of a sparsely activated model, we further define sparsity ratio as $ SR = n_{expert} / k $. Model with higher sparsity ratio is sparser.


\noindent\textbf{Proof} For the MoE layer, the computational cost is proportional to k:

\begin{equation}\label{cost}
    {\rm cost}_{compute} \propto k \cdot d_{model} \cdot d_{ffn},
\end{equation}
and the relation of $ s_{model} $ and $ n_{expert} $ (ignore the size of router) is:

\begin{equation}\label{size}
    s_{model} = n_{expert} \cdot d_{model} \cdot d_{ffn}.
\end{equation}

According to Eq. (\ref{cost}), Eq. (\ref{size}) and the definition of sparsity ratio, we have:

\begin{equation}
     s_{model} \propto SR \cdot {\rm cost}_{compute}.
\end{equation}

Therefore, we can conclude that the model size is only proportional to the sparsity ratio at a certain computational cost.

First, we conduct experiments on the sparsely activated models with $ k=1 $ using Switch routing~\cite{william-2021-swt}. As shown in Figure \ref{sr_loss}, we found that when the sparsity ratio of the model is 32, the performance of the model is optimal. With the sparsity ratio further increased, the capacity of the model is enlarged, but the performance of the model cannot be improved anymore. Therefore, with single activated expert, when the sparsity ratio is 32, the performance of the model reaches its optimal, which we call \textit{sparsity ceiling}.

As shown in the Figure \ref{sr_loss}, when $ k=4 $, the performance of the model becomes saturated when the sparsity ratio is 128. That is, when we activate four experts, the sparsity ratio with which the model reaches the best performance is larger than activating only one expert, and the performance of the model is better, too. By further increasing $k$, we can see that the sparsity ratio reaching the best performance of model will become larger and larger, and its performance will also be improved. In other words, increasing $k$ value will not only enhance the performance of the model, but also enhance the scalability of the model.

In summary, increasing $k$ not only breaks through the performance ceiling of the model, but also improves the scalability, and the performance of larger models can be fully utilized. However, due to the current hardware limitations, increasing $k$ will add unacceptable communication cost into the model calculation process, which seriously affects the model efficiency. To address this issue, in next section, we propose a brand new routing mechanism that can reduce the communication cost and increase the performance of the model at the same time.

\section{Switch and Mixture Routing}

To achieve extreme sparsity of sparsely activated models with low communication cost, we propose a Switch and Mixture layer (SAM layer), to improve the performance and scalability with low communication cost. The main architecture of SAM layer is shown in Figure \ref{sam_architecture}. In this section, we will introduce the SAM layer in three parts: hierarchical routing, alignment loss and load balance.

\subsection{Hierarchical Routing}

As we have analyzed in Section \ref{topk}, in order to boost the performance of large-scale sparse models, increasing $k$ (the number of activated experts) is a necessary choice. Meanwhile, such huge models require computations in distributed systems with many devices(usually GPUs). However, as $k$ increases, the cost of communication between devices gradually becomes unacceptable. To address this, we propose a novel hierarchical routing mechanism which is called Switch and Mixture routing (SAM). Specifically, we group experts according to their location, which means experts on the same device are in the same group. During the forward computation, we first select one group and then activate multiple experts located in this group(also means in the same device). Obviously, with the SAM routing, the communication cost will be decoupled with the number of activated experts, thus the efficiency is guaranteed.

In SAM routing as shown in Figure \ref{sam_architecture}, the first router (called Switch Router) is used to select one group of experts, while the second (called Mixture Router) is used to select $k$ experts in the selected group.

Firstly, similarly with recent works~\cite{dmitry-2020-gshard,william-2021-swt}, we use Shared Router to implement it.


\subsubsection{Shared Router}

In Shared Router implementation, a router is shared when selecting groups and experts. It computes scores of experts globally:

\begin{equation}
    p(x)_i =  {{\rm softmax}(w_{s} \cdot h_x)
    }_i,
\end{equation}
where $w_{s}$ is the learnable parameters of the shared router, $h_x$ is the hidden state of token $x$ and $p(x)_i$ is the score of $i$-th expert. Accordingly, the score for each group will be calculated as the sum of Top-k scores in the group :

\begin{equation}
    g(x)_w = \sum_{i \in \mathcal{T}_w}p(x)_i.
\end{equation}
where $ \mathcal{T}_w $ is the set of Top-k experts in $w$-th group. $g(x)_w$ is the score of $w$-th group and according this score, router will select the best group to compute. In selected group, we select the experts with Top-k scores given by the Shared router and use $ \mathcal{T} $ to represent this set, the final result will be calculate as:

\begin{equation}
    y = \sum_{i \in \mathcal{T}}p(x)_i E(x)_i.
\end{equation}





However, using Shared Router can not represent the diversity between groups efficiently, further, we propose non-shared router to improve this.

\subsubsection{Non-shared Router}

In Non-shared Router implementation, we propose Switch Router which calculate the scores of groups and Mixture Router calculate the scores of experts. As for Switch Router, the score of $w$-th group calculated as:

\begin{equation}
    g(x)_w =  {\rm softmax}(w_{ns} \cdot h_x)_w,
\end{equation}
where $w_{ns}$ is the parameters of the Switch Router of Non-shared Router and $g(x)_w$ is the score of $w$-th group.

After selecting the best group according Switch Router, Mixture Router calculate the score of $i$-th expert as:

\begin{equation}
    p_{w}(x)_i =  {\rm softmax}(w_{nm}^{w} \cdot h_x)_i,
    \label{mixture_router}
\end{equation}
where $w_{nm}^{w}$ is the parameters of the Mixture Router of $w$-th group and $p_{w}(x)_i$ is the score of $i$-th expert which located in $w$-th group. Each group have independent Mixture Router, and each Mixture Router calculate the scores inner experts of group.

Finally each expert in the selected group has two scores: one is the score of the selected group ($ g(x)_w $) and the other is of the selected expert ($ p_{w}(x)_i $). The output is given by:

\begin{equation}
    y = \sum_{i \in \mathcal{T}} g(x)_w \cdot p_{w}(x)_i \cdot E_{i}^{w}(h_x),
\end{equation}
where $w$-th group is selected group, $ g(x)_w $ is the score of this group, $ p_{w}(x)_i $ is score of $i$-th expert located in $w$-th group, $ \mathcal{T} $ is set of the selected experts, and $ E_{i}^{w}(h_x) $ is the output of the $i$-th expert in $w$-th group.




\subsection{Alignment Loss}

With hierarchical routing, especially Shared Router, we find that the Top-k scores of experts in the selected group may not be the best Top-k in global. So we design a loss to align the best score of local with global:

\begin{equation}
    loss = \sum_{e \in \mathcal{T}}\max\left((p(x)_e - p(x)_{K}),0\right),
    \label{group_loss}
\end{equation}
where $ \mathcal{T} $ is the set of experts not located in the group that we selected and $ p(x)_{K} $ is the $k$-th highest score of the selected experts.
Using this loss restricts the scores of other experts lower than selected experts.

Relatively, as for non-shared router, we also design a loss function to restrict results of routers:


\begin{equation}
    loss = {\rm -log} \frac{g(x)_k}{\mathop{\sum \limits_{w \in \mathcal{W}}} g(x)_w},
    \label{log_loss}
\end{equation}
where $ g(x)_k $ means the score of selected group and $\mathcal{W}$ is the set of groups.




\subsection{Load Balance}


In the sparsely activated models, due to activating experts sparsely for calculation, the load imbalance of tokens may occur when tokens are allocated. When the load imbalance occurs, a large proportion of the tokens will be routed to a small number of experts, leading to inadequate training of other experts. As a result, the performance will be deteriorated.

 In Switch Router, we use auxiliary loss like~\cite{william-2021-swt} to keep the load balance between groups and in Mixture Router, we use it to keep balance between experts.

\begin{figure*}
	\centering
	\setlength{\belowcaptionskip}{-0.1cm}
	\subfigure[Switch-GPT]{
		\begin{minipage}[b]{0.3\textwidth}
		    \includegraphics[width=1.1\textwidth]{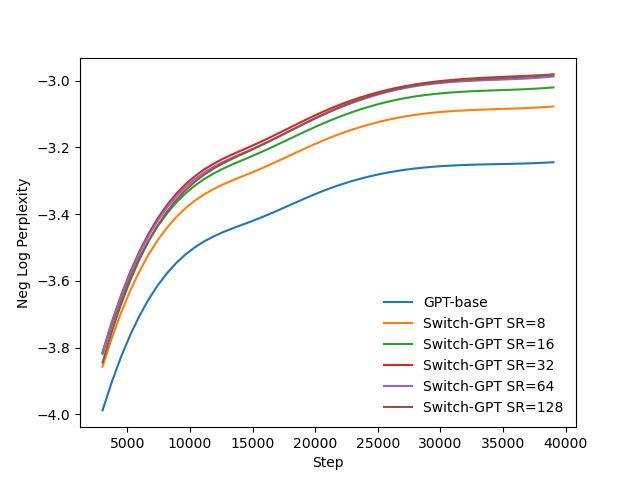}
		\end{minipage}
		\label{switch bottleneck}
	}
	\subfigure[SAM-GPT (k=2)]{
		\begin{minipage}[b]{0.3\textwidth}
		    \includegraphics[width=1.1\textwidth]{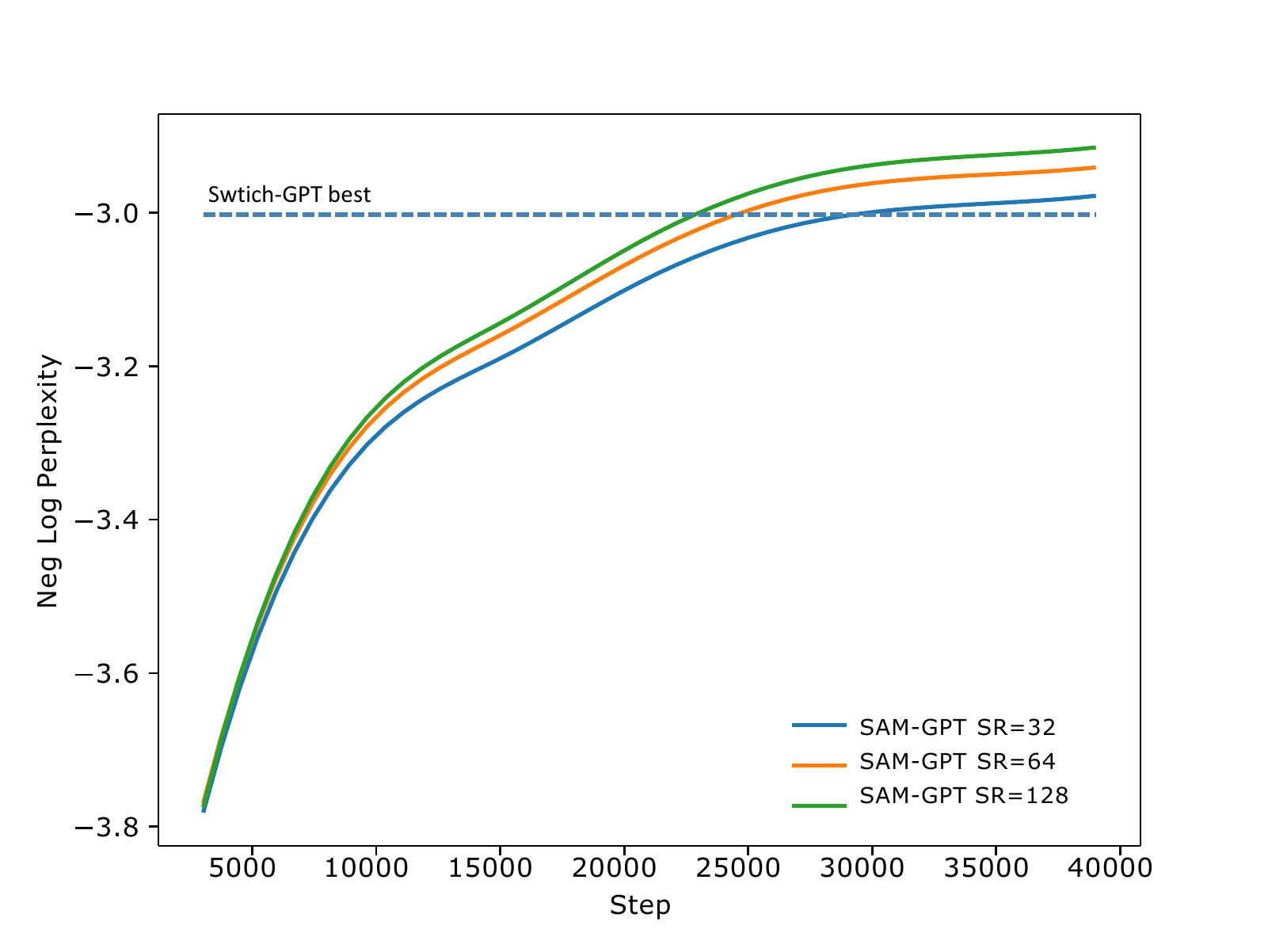}
		\end{minipage}
		\label{sam k2 bottleneck}
	}
	\subfigure[SAM-GPT (k=4)]{
		\begin{minipage}[b]{0.3\textwidth}
  	 	    \includegraphics[width=1.1\textwidth]{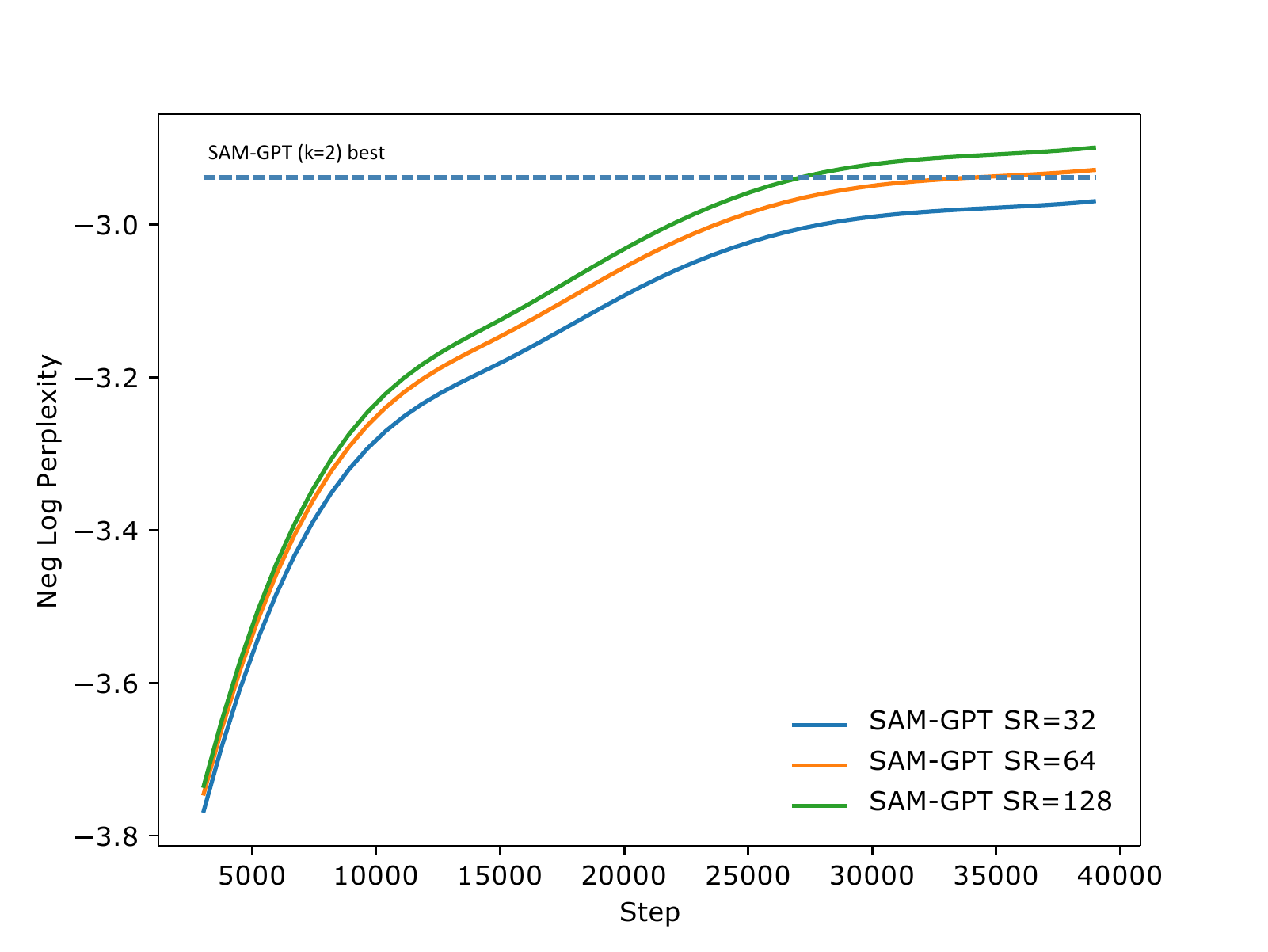}
		\end{minipage}
	    \label{sam k4 bottleneck}
	}
	\caption{\textbf{Performance of different sparsity ratio}. Figure \ref{switch bottleneck} shows the performance of Switch-GPT, model will reach the best performance when sparsity ratio is 32 and increase sparsity will not improve the performance anymore  Figure \ref{sam k2 bottleneck} and Figure \ref{sam k4 bottleneck} shows SAM-GPT, we can increase the sparsity of model to increase the performance continuously.}
	\label{bottleneck}
\end{figure*}

\section{Experiments}




\subsection{Experimental Details}

In this work, we use GPT-base~\cite{Radford2018ImprovingLU} as the backbone model. We replace all the FFN layers in GPT-base with Switch routing and SAM routing, and obtain two kinds of sparsely activated models called Switch-GPT and SAM-GPT, respectively.

Following the original settings in ~\cite{Radford2018ImprovingLU}, the model size of all the discussed models is set to 768. The hidden size of FFN layers in GPT-base and Switch-GPT (always activates a single FFN expert at a time) is 3072 (4 times model size). In order to keep computational cost of all the models consistent, the hidden size of SAM-GPT is 3072/k, where k is the number of activated FFN experts in sparse routing. All the models are optimized by the Adam~\cite{Kingma-2015-adam} optimizer with batch size 128.

In order to explore the ability of the model, we test the perplexity for the language model in the pre-training stage. The pre-training corpus will be introduced in the next section. Section \ref{ceiling} and Section \ref{efficiency} respectively show experiments to prove performance and efficiency improvement. At the same time, in order to explore the model's ability to model long-distance in natural language and measure the amount of knowledge, we carry out some downstream tasks in Section \ref{downstream}.

\subsection{Pre-training Dataset}

To prevent inadequate training due to the insufficiency of data, we follow the work of ~\cite{shoebi-2019-megatron} and aggregate three datasets for pre-training, including Wikipedia~\cite{devlin-2019-bert}, RealNews~\cite{zellers-2019-realnews} and OpenWebText~\cite{radford-2019-gpt2}. Wikipedia is a dataset composed of extracted texts from wikipedia without lists, tables and headers. RealNews is a large scale dataset about news articles from Common Crawl. OpenWebText dataset contains a large amount of text data from social media platform Reddit. They scraped all outbound links from Reddit as data sources and extracted texts from HTML responses. To avoid data leakage, we deleted the data in WikiText-103. All together, the combined dataset contains nearly 165GB data.

\subsection{Breaking though the Ceiling of Sparsity}\label{ceiling}

In this section, we conduct experiments to show that using SAM layer we can break through the ceiling of sparsity described in Section \ref{topk}.

As shown in Figure \ref{switch bottleneck}, Switch-GPT which activated a single expert, will get better result than the dense model  with the same computational cost. As the number of experts increases, the performance becomes better. When sparsity ratio is 32, Switch-GPT gets the best result. However, as the sparsity increases continuously, the performance of the model does not improve accordingly. In other words, the Switch-GPT model reaches the Ceiling of Sparsity when the sparsity ratio is 32.

\begin{table*}
    \centering
    \begin{tabular}{lllllll}
    \toprule
        Model & Parameters & Sparsity Ratio & Number of Expert  & $d_{model}$ & $d_{FFN}$ & Perplexity \\ \hline
        GPT-base & 110M & - & - & 768 & 3072 & 25.89 \\
        Switch-GPT & 3.6B & 64 & 64 & 768 & 3072 &  20.26 \\ \hline
        SAM-GPT (k=2) & 3.6B & 64 & 128 & 768 & 1536 & 19.39 \\
        SAM-GPT (k=4) & 3.6B & 64 & 256 & 768 & 768 & 19.17 \\
        SAM-GPT (k=2) best & 7.2B & 128 & 256 & 768 & 1536 & 18.88  \\
        SAM-GPT (k=4) best & 7.2B & 128 & 512 & 768 & 768 &  \textbf{18.53}  \\ \toprule
    \end{tabular}
    \caption{\textbf{Benchmarking SAM-GPT, Switch-GPT and GPT.} In this table, we compare the performance of GPT-base, Switch-GPT and SAM-GPT with same computational cost. SAM-GPT achieves better performance compared with Switch-GPT and dense model (GPT-base). SAM-GPT with k = 2 reaches the best performance when the sparsity ratio is 128 and corresponding perplexity is 18.88. When the value of k is further increased to 4, the best performance will be further improved accordingly with a lower perplexity of 18.53.}
    \label{result_1}
\end{table*}

In contrast, Figure \ref{sam k2 bottleneck} and Figure \ref{sam k4 bottleneck} show the performance comparison of SAM routing with different sparsity ratios at different $k$ values.
Figure \ref{sam k2 bottleneck} shows the performance of SAM layer which activates 2 experts. Compared with dense model (GPT-base) with the same computational cost, SAM-GPT has a great improvement even when sparsity ratio is 32. As the sparsity ratio increases further, the model performs better and better. When sparsity ratio is 128, it takes SAM-GPT only 22k steps to achieve the performance of Switch-GPT. In addition, compared with sparsity ratio is 64, the model is sparser and the performance is much better. In summary, the experiments show that when two experts are activated, not only is the performance of the SAM-GPT significantly better than Switch-GPT, but its sparsity is much higher, too.

By further increasing the number of activated experts, in Figure \ref{sam k4 bottleneck}, the improvement will be much better than activating only two experts. It is reasonable to believe that increasing the number of activated experts through SAM-GPT extends the performance upper bound and extensibility of the model to some extent. Compared with other work, we can activate more experts to improve the model's performance and train efficient sparsely activated models based on this.


The overall results are shown in Table \ref{result_1}. With the same computational cost, the performance of sparsely activated models is better than dense model. The perplexity of GPT-base is 25.89 and Switch-GPT and SAM-GPT are both always better than it. In the case of under the same communication cost and calculation cost, SAM-GPT can activate multiple experts for calculation, and its effect will be much better than Switch-GPT. With the same sparsity ratio, activating 2 experts reaches 19.39 PPL, and activating 4 gets 19.17. Further increase sparsity ratio, the performance will be better. As the table shows, SAM-GPT (k=2) will get the best performance when sparsity ratio is 128 and PPL is 18.88, SAM-GPT (k=4) will get the best result is 18.53.

\begin{figure}
    \centering
    \includegraphics[width=0.45\textwidth]{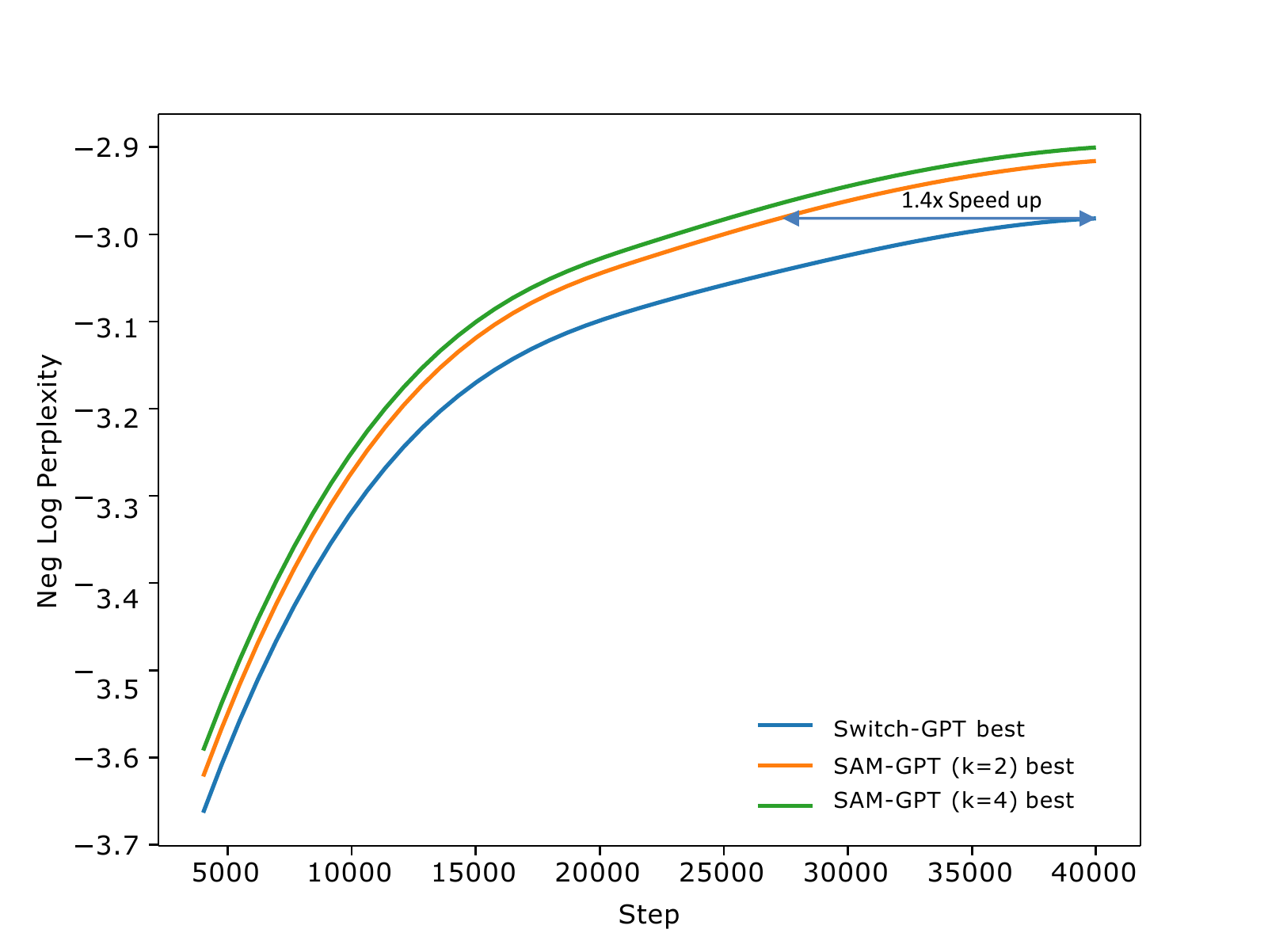}
    \caption{\textbf{The performance of Switch-GPT, SAM-GPT (k=2) and SAM-GPT (k=4) in same sparsity ratio} In this Figure, the computational cost is same, and all model's sparsity ratio is 128. We can see the result of SAM-GPT (k=4) is better than other models and get 1.4x speedup in training compared with Switch-GPT.}
    \label{same_sr128}
\end{figure}

\subsection{Efficiency of SAM Transformer}\label{efficiency}

Compared with Switch-GPT, SAM-GPT has a significant improvement not only in performance but also in efficiency. In Figure \ref{same_sr128}, we can observe that the performance of SAM-GPT is better than Switch-GPT, even when activating only 2 experts. Furthermore, at the same step, SAM-GPT will be better than Switch-GPT. Unlike dense model, the performance improvement due to capacity growth by increasing the sparsity does not
give rise to the increase of computational cost. As shown in the Figure \ref{same_sr128}, it takes SAM-GPT only 20k training steps to achieve the best performance of Switch-GPT. According to this, we can use the SAM routing to increase the number of activated experts and sparsity of the model and then improve the performance. SAM-GPT deceases the communication cost by expert grouping, however, comparing steps cannot show the efficiency of the SAM routing clearly. Further, we will compare Switch routing and MoE routing in total elapsed time and show the efficiency of the SAM routing. General speaking, the SAM routing gives a choice to build such a huge and economic model for natural language processing.

\subsection{Downstream Task}\label{downstream}

\begin{table}
    \centering
    \begin{tabular}{llll}
    \toprule
        \multirow{2}*{Model} & LAMB. & Wiki. & LAMA   \\
        & (Acc.) & (PPL) & (P@1)\\ \hline
        GPT-base & 0.2304 & 40.57 & 0.1084   \\
        Switch-GPT & 0.2581 & 38.46 & 0.1365 \\
        SAM-GPT & \textbf{0.2806} & \textbf{28.48} & \textbf{0.1549} \\ \toprule
    \end{tabular}
    \caption{\textbf{Zero-shot results on LAMBADA (LAMB.), WikiText-103 (Wiki.) and LAMA.}}
    \label{downstream_z}
\end{table}



In this section, we conduct some experiments to evaluate the performance of SAM-GPT in some downstream tasks about neural language generation and modeling. 

In Table \ref{downstream_z}, We choose LAMBADA~\cite{paperno-etal-2016-lambada},  WikiText-103~\cite{merity-2016-wikitext103} and  LAMA~\cite{fabio-2019-lama} to test the ability of models to model long-term dependencies in text and report the results of GPT-base, Switch-GPT and SAM-GPT. The metric of LAMBADA and LAMA is accuracy and WikiText-103 is perplexity. As shown in Table \ref{downstream_z}, SAM-GPT outperforms Switch-GPT nearly 2.2 points in LAMBDA, 10 PPL in WikiText-103 and 2 points in LAMA. SAM-GPT has the best result in such knowledge-heavy tasks.

\section{Related Work}

Pre-training has become a focus in the field of artificial intelligence, especially in natural language processing. After Transformer~\cite{ashish-2017-transformer} is proposed, pre-training has become a common choice in natural language processing, and enhancing downstream tasks through pre-training has become the mainstream method.~\cite{devlin-2019-bert, yihan-2019-roberta, mike-2019-bart, radford-2019-gpt2, tom-2020-gpt3} For example,~\cite{devlin-2019-bert} proposed BERT which constructed Bidirectional Encoder based on self-attention mechanism, and trained it as a masked language model, which greatly improved the modeling and understanding ability of the language model.

~\cite{jared-2020-scale} have explored various factors that affect the neural network and found that model size is a significant factor to affect model performance. However, if we increase the number of layers or the size of hidden layers, training instability and gradient explosion may occur during training. Besides, increasing the model size of a dense model will greatly increase the computational cost of the model. Inspired by the Mixture of Experts~\cite{jacobs-91-moe, jordan-93-moe, noam-2017-moe}, some researchers begin to study sparsely activated models to improve pre-trained models based on self-attention mechanism. In recent years, with the rapid development of distributed training and parallel computing, a large amount of large-scale pre-trained language model research has been released.~\cite{dmitry-2020-gshard} expanded their FFN layers based on Transformer and acquired SOTA on translation tasks of various languages.~\cite{william-2021-swt} propose Switch Transformer, which simplifies the routing of MoE, reduces communication cost, and expands the scale to trillion. Although previous studies have scaled the model to trillion, the benefits of performance are not as good. This work improves the performance bottleneck of scaling up the model, and reduces the communication overhead through hierarchical routing to achieve a truly efficient large-scale language model.

\section{Conclusion and Future Work}

In this work, we experiment several factors related to the performance of sparsely activated models, and find that when increasing the number of experts, the improvement of performance will reach a ceiling. Continuing to increase the sparsity of model will not get the benefit of performance. Further more, we find that increasing the number of activated experts can effectively improve the performance and break through the ceiling of sparsity. However, due to hardware limitations, activating multiple experts can always incur unacceptable communication cost.

Therefore, we propose SAM routing to select experts by hierarchical routing. The Switch router selects a device and the Mixture router chooses experts in the selected device. This decouples the communication cost with the number of activated experts to reduce communication cost. Experiments show that our method is significantly better than Switch routing in terms of both performance and efficiency.


At present, most sparsely activated models based on Transformer choose to expand the FFN layer to improve the performance of expanding model capacity. We think that not only can we expand the FFN layer, but also on the self-attention layer. Further, we can expand the whole Transformer layer or the whole model to store and model knowledge and release the ability of understanding. In addition, there is still a certain gap between SAM routing and MoE routing in the final effect. We consider that the diversity of experts may be restricted due to grouping, which remains as an important research direction in our future work.

\bibliography{custom}

\begin{thebibliography}{22}
\expandafter\ifx\csname natexlab\endcsname\relax\def\natexlab#1{#1}\fi

\bibitem[{Brown et~al.(2020)Brown, Mann, Ryder, Subbiah, Kaplan, Dhariwal,
  Neelakantan, Shyam, Sastry, Askell, Agarwal, Herbert{-}Voss, Krueger,
  Henighan, Child, Ramesh, Ziegler, Wu, Winter, Hesse, Chen, Sigler, Litwin,
  Gray, Chess, Clark, Berner, McCandlish, Radford, Sutskever, and
  Amodei}]{tom-2020-gpt3}
Tom~B. Brown, Benjamin Mann, Nick Ryder, Melanie Subbiah, Jared Kaplan,
  Prafulla Dhariwal, Arvind Neelakantan, Pranav Shyam, Girish Sastry, Amanda
  Askell, Sandhini Agarwal, Ariel Herbert{-}Voss, Gretchen Krueger, Tom
  Henighan, Rewon Child, Aditya Ramesh, Daniel~M. Ziegler, Jeffrey Wu, Clemens
  Winter, Christopher Hesse, Mark Chen, Eric Sigler, Mateusz Litwin, Scott
  Gray, Benjamin Chess, Jack Clark, Christopher Berner, Sam McCandlish, Alec
  Radford, Ilya Sutskever, and Dario Amodei. 2020.
\newblock \href {http://arxiv.org/abs/2005.14165} {Language models are few-shot
  learners}.
\newblock \emph{CoRR}, abs/2005.14165.

\bibitem[{Devlin et~al.(2018)Devlin, Chang, Lee, and
  Toutanova}]{devlin-2019-bert}
Jacob Devlin, Ming{-}Wei Chang, Kenton Lee, and Kristina Toutanova. 2018.
\newblock \href {http://arxiv.org/abs/1810.04805} {{BERT:} pre-training of deep
  bidirectional transformers for language understanding}.
\newblock \emph{CoRR}, abs/1810.04805.

\bibitem[{Fedus et~al.(2021)Fedus, Zoph, and Shazeer}]{william-2021-swt}
William Fedus, Barret Zoph, and Noam Shazeer. 2021.
\newblock \href {http://arxiv.org/abs/2101.03961} {Switch transformers: Scaling
  to trillion parameter models with simple and efficient sparsity}.
\newblock \emph{CoRR}, abs/2101.03961.

\bibitem[{Hochreiter and Schmidhuber(1997)}]{Ho-97-lstm}
Sepp Hochreiter and J\"{u}rgen Schmidhuber. 1997.
\newblock \href {https://doi.org/10.1162/neco.1997.9.8.1735} {Long short-term
  memory}.
\newblock \emph{Neural Comput.}, 9(8):1735–1780.

\bibitem[{Jacobs et~al.(1991)Jacobs, Jordan, Nowlan, and
  Hinton}]{jacobs-91-moe}
Robert~A. Jacobs, Michael~I. Jordan, Steven~J. Nowlan, and Geoffrey~E. Hinton.
  1991.
\newblock \href {https://doi.org/10.1162/neco.1991.3.1.79} {Adaptive mixtures
  of local experts}.
\newblock \emph{Neural Computation}, 3(1):79--87.

\bibitem[{Jordan and Jacobs(1993)}]{jordan-93-moe}
M.I. Jordan and R.A. Jacobs. 1993.
\newblock \href {https://doi.org/10.1109/IJCNN.1993.716791} {Hierarchical
  mixtures of experts and the em algorithm}.
\newblock In \emph{Proceedings of 1993 International Conference on Neural
  Networks (IJCNN-93-Nagoya, Japan)}, volume~2, pages 1339--1344 vol.2.

\bibitem[{Kaplan et~al.(2020)Kaplan, McCandlish, Henighan, Brown, Chess, Child,
  Gray, Radford, Wu, and Amodei}]{jared-2020-scale}
Jared Kaplan, Sam McCandlish, Tom Henighan, Tom~B. Brown, Benjamin Chess, Rewon
  Child, Scott Gray, Alec Radford, Jeffrey Wu, and Dario Amodei. 2020.
\newblock \href {http://arxiv.org/abs/2001.08361} {Scaling laws for neural
  language models}.
\newblock \emph{CoRR}, abs/2001.08361.

\bibitem[{Kingma and Ba(2015)}]{Kingma-2015-adam}
Diederik~P. Kingma and Jimmy Ba. 2015.
\newblock Adam: A method for stochastic optimization.
\newblock \emph{CoRR}, abs/1412.6980.

\bibitem[{Lepikhin et~al.(2020)Lepikhin, Lee, Xu, Chen, Firat, Huang, Krikun,
  Shazeer, and Chen}]{dmitry-2020-gshard}
Dmitry Lepikhin, HyoukJoong Lee, Yuanzhong Xu, Dehao Chen, Orhan Firat, Yanping
  Huang, Maxim Krikun, Noam Shazeer, and Zhifeng Chen. 2020.
\newblock \href {http://arxiv.org/abs/2006.16668} {Gshard: Scaling giant models
  with conditional computation and automatic sharding}.
\newblock \emph{CoRR}, abs/2006.16668.

\bibitem[{Lewis et~al.(2019)Lewis, Liu, Goyal, Ghazvininejad, Mohamed, Levy,
  Stoyanov, and Zettlemoyer}]{mike-2019-bart}
Mike Lewis, Yinhan Liu, Naman Goyal, Marjan Ghazvininejad, Abdelrahman Mohamed,
  Omer Levy, Veselin Stoyanov, and Luke Zettlemoyer. 2019.
\newblock \href {http://arxiv.org/abs/1910.13461} {{BART:} denoising
  sequence-to-sequence pre-training for natural language generation,
  translation, and comprehension}.
\newblock \emph{CoRR}, abs/1910.13461.

\bibitem[{Liu et~al.(2019)Liu, Ott, Goyal, Du, Joshi, Chen, Levy, Lewis,
  Zettlemoyer, and Stoyanov}]{yihan-2019-roberta}
Yinhan Liu, Myle Ott, Naman Goyal, Jingfei Du, Mandar Joshi, Danqi Chen, Omer
  Levy, Mike Lewis, Luke Zettlemoyer, and Veselin Stoyanov. 2019.
\newblock \href {http://arxiv.org/abs/1907.11692} {Roberta: {A} robustly
  optimized {BERT} pretraining approach}.
\newblock \emph{CoRR}, abs/1907.11692.

\bibitem[{Merity et~al.(2016)Merity, Xiong, Bradbury, and
  Socher}]{merity-2016-wikitext103}
Stephen Merity, Caiming Xiong, James Bradbury, and Richard Socher. 2016.
\newblock \href {http://arxiv.org/abs/1609.07843} {Pointer sentinel mixture
  models}.
\newblock \emph{CoRR}, abs/1609.07843.

\bibitem[{Narayanan et~al.(2021)Narayanan, Shoeybi, Casper, LeGresley, Patwary,
  Korthikanti, Vainbrand, Kashinkunti, Bernauer, Catanzaro, Phanishayee, and
  Zaharia}]{deepak-21-megathronlm}
Deepak Narayanan, Mohammad Shoeybi, Jared Casper, Patrick LeGresley, Mostofa
  Patwary, Vijay Korthikanti, Dmitri Vainbrand, Prethvi Kashinkunti, Julie
  Bernauer, Bryan Catanzaro, Amar Phanishayee, and Matei Zaharia. 2021.
\newblock \href {http://arxiv.org/abs/2104.04473} {Efficient large-scale
  language model training on {GPU} clusters}.
\newblock \emph{CoRR}, abs/2104.04473.

\bibitem[{Paperno et~al.(2016)Paperno, Kruszewski, Lazaridou, Pham, Bernardi,
  Pezzelle, Baroni, Boleda, and Fern{\'a}ndez}]{paperno-etal-2016-lambada}
Denis Paperno, Germ{\'a}n Kruszewski, Angeliki Lazaridou, Ngoc~Quan Pham,
  Raffaella Bernardi, Sandro Pezzelle, Marco Baroni, Gemma Boleda, and Raquel
  Fern{\'a}ndez. 2016.
\newblock \href {https://doi.org/10.18653/v1/P16-1144} {The {LAMBADA} dataset:
  Word prediction requiring a broad discourse context}.
\newblock In \emph{Proceedings of the 54th Annual Meeting of the Association
  for Computational Linguistics (Volume 1: Long Papers)}, pages 1525--1534,
  Berlin, Germany. Association for Computational Linguistics.

\bibitem[{Petroni et~al.(2019)Petroni, Rockt{\"{a}}schel, Lewis, Bakhtin, Wu,
  Miller, and Riedel}]{fabio-2019-lama}
Fabio Petroni, Tim Rockt{\"{a}}schel, Patrick S.~H. Lewis, Anton Bakhtin,
  Yuxiang Wu, Alexander~H. Miller, and Sebastian Riedel. 2019.
\newblock \href {http://arxiv.org/abs/1909.01066} {Language models as knowledge
  bases?}
\newblock \emph{CoRR}, abs/1909.01066.

\bibitem[{Radford and Narasimhan(2018)}]{Radford2018ImprovingLU}
Alec Radford and Karthik Narasimhan. 2018.
\newblock Improving language understanding by generative pre-training.

\bibitem[{Radford et~al.(2019)Radford, Wu, Child, Luan, Amodei, Sutskever
  et~al.}]{radford-2019-gpt2}
Alec Radford, Jeffrey Wu, Rewon Child, David Luan, Dario Amodei, Ilya
  Sutskever, et~al. 2019.
\newblock Language models are unsupervised multitask learners.
\newblock \emph{OpenAI blog}, 1(8):9.

\bibitem[{Raffel et~al.(2019)Raffel, Shazeer, Roberts, Lee, Narang, Matena,
  Zhou, Li, and Liu}]{colin-2019-t5}
Colin Raffel, Noam Shazeer, Adam Roberts, Katherine Lee, Sharan Narang, Michael
  Matena, Yanqi Zhou, Wei Li, and Peter~J. Liu. 2019.
\newblock \href {http://arxiv.org/abs/1910.10683} {Exploring the limits of
  transfer learning with a unified text-to-text transformer}.
\newblock \emph{CoRR}, abs/1910.10683.

\bibitem[{Shazeer et~al.(2017)Shazeer, Mirhoseini, Maziarz, Davis, Le, Hinton,
  and Dean}]{noam-2017-moe}
Noam Shazeer, Azalia Mirhoseini, Krzysztof Maziarz, Andy Davis, Quoc~V. Le,
  Geoffrey~E. Hinton, and Jeff Dean. 2017.
\newblock \href {http://arxiv.org/abs/1701.06538} {Outrageously large neural
  networks: The sparsely-gated mixture-of-experts layer}.
\newblock \emph{CoRR}, abs/1701.06538.

\bibitem[{Shoeybi et~al.(2019)Shoeybi, Patwary, Puri, LeGresley, Casper, and
  Catanzaro}]{shoebi-2019-megatron}
Mohammad Shoeybi, Mostofa Patwary, Raul Puri, Patrick LeGresley, Jared Casper,
  and Bryan Catanzaro. 2019.
\newblock \href {http://arxiv.org/abs/1909.08053} {Megatron-lm: Training
  multi-billion parameter language models using model parallelism}.
\newblock \emph{CoRR}, abs/1909.08053.

\bibitem[{Vaswani et~al.(2017)Vaswani, Shazeer, Parmar, Uszkoreit, Jones,
  Gomez, Kaiser, and Polosukhin}]{ashish-2017-transformer}
Ashish Vaswani, Noam Shazeer, Niki Parmar, Jakob Uszkoreit, Llion Jones,
  Aidan~N. Gomez, Lukasz Kaiser, and Illia Polosukhin. 2017.
\newblock \href {http://arxiv.org/abs/1706.03762} {Attention is all you need}.
\newblock \emph{CoRR}, abs/1706.03762.

\bibitem[{Zellers et~al.(2019)Zellers, Holtzman, Rashkin, Bisk, Farhadi,
  Roesner, and Choi}]{zellers-2019-realnews}
Rowan Zellers, Ari Holtzman, Hannah Rashkin, Yonatan Bisk, Ali Farhadi,
  Franziska Roesner, and Yejin Choi. 2019.
\newblock \href {http://arxiv.org/abs/1905.12616} {Defending against neural
  fake news}.
\newblock \emph{CoRR}, abs/1905.12616.

\end{thebibliography}
\bibliographystyle{acl_natbib}

\appendix



\end{document}